# Evaluating GPT-3.5's Awareness and Summarization Abilities for European Constitutional Texts with Shared Topics

Candida M. Greco and Andrea Tagarelli

DIMES Dept., University of Calabria, 87036 Rende, Italy
{candida.greco,tagarelli}@dimes.unical.it
https://mlnteam-unical.github.io/

**Abstract.** Constitutions are foundational legal documents that underpin the governmental and societal structures. As such, they are a reflection of a nation's cultural and social uniqueness, but also contribute to establish topics of universal importance, like citizens' rights and duties (RD). In this work, using the renowned GPT-3.5, we leverage generative large language models to understand constitutional passages that transcend national boundaries. A key contribution of our study is the introduction of a novel application of abstractive summarization on a multi-source collection of constitutional texts, with a focus on European countries' constitution passages related to RD topics. Our results show the meaningfulness of GPT-3.5 to produce informative, coherent and faithful summaries capturing RD topics across European countries.

## 1 Introduction

Constitutions encapsulate the fundamental principles and rights upon which societies are established. When constitutions concern sovereign countries, they can be regarded as a proxy to reflect the unique historical, cultural and social contexts of their respective nations. Yet, amid this diversity, constitutions contain a backbone of themes of universal importance, such as core principles of democracy and human rights. In this respect, our research interest is in understanding and unveiling shared constitutional elements among countries that transcend national boundaries. We believe this holds profound practical and academic implications. First, it can enable policymakers and legal professionals to more deeply and quickly identify areas of agreement, reducing the time and effort required for decision-making and policy formulation. Second, in a federation of countries or a supranational political/economical union of states, it can promote legal alignment, efficiency and transparency among countries, making a fundamental step towards a more integrated and cooperative legal landscape. In this regard, the study in [2] conducted a comparative analysis of constitutions from the US and other world countries to identify similarities in citizens' rights and government institution relations.

In this work, we aim to take a first step towards the aforementioned research problem through learning a summarization of European constitutional texts that share a number of topics that are relevant to the realm of *rights and duties* of citizens, hereinafter referred to as *RD-relevant topics*. Our rationale for this focus is twofold. First, Europe per se represents a highly relevant case in point, telling a rich and centuries-old history involving all European populations; moreover, there has been an increased interest in processing European law documents, thanks also to a publicly available access to vast resources of the European Union [3]. Second, rights and duties are at the core of a society, shaping the relationship between individuals and the state, thus representing a thematic framework that establishes the boundaries of governmental authority, safeguards individual liberties, and ensures a just and equitable society.

Manual summarization methods would clearly struggle to cope with the vastness and complexity of constitutional texts, thus opening opportunities for an exploration of automated alternatives. Indeed, the use of AI has long been a topic of interest in the law area [15, 12, 9]. In this regard, following the wave triggered by generative AI, we want to leverage the reasoning and summarization capabilities of generative large language models (LLMs) from the decoder-only autoregressive family, a subset of the broader category of LLMs. Since the momentum gained after the introduction of ChatGPT (chat.openai.com), these models have demonstrated remarkable proficiency in generating coherent natural language responses when provided with appropriate prompts, being able to interpret the instructions specified therein and execute them effectively [8]. Legal actors, such as law professionals and firms, have expressed interest in exploiting these models to enhance legal search and document review tasks, achieving scalability and cost-efficiency. The adoption of LLMs in legal AI is an ever-expanding trend [13, 6], and it largely encompasses methods for summarization. Based on the above premises, our study lays on the following research questions:

**RQ1:** *Does a GPT-like LLM currently have adequate knowledge of the European constitutions? If so, what are the RD-relevant topics that the generative LLM is acquainted with?*

**RQ2:** *If prompted with portions of country-specific constitutions about RD-relevant topics, how well is a generative LLM able to produce a coherent summary capturing the common constitutional law across all countries while keeping the main distinctive country-specific topical signals?*

Our **contributions** can be summarized as follows. **(1)** To the best of our knowledge, we are the first to address the above research questions by leveraging a generative LLM. Although the current panorama of such models is not limited to a single solution, both within the GPT family and other commercially-licensed models as well as open source models, in this work we choose to experiment with `gpt-3.5-turbo` (4k and 16k context-window versions),[1] which is considered to be the most groundbreaking and popular generative AI for NLP, still under the spotlight in many contexts. **(2)** We introduce a novel application of abstractive summarization on a multi-source collection of texts from the constitutions of European countries dealing with RD-relevant topics. This is a challenging testbed, given not only the intricacies and subtleties of the legal sublanguage but also because of the intrinsic complexity of dealing with foundational laws of different countries which are thematically related yet diverse in many different aspects concerning their history, culture, ethnicity, religion, and more. **(3)** We conducted a comprehensive evaluation of GPT-3.5 to pursue two goals. First, we assessed the degree of its awareness of the topics covered in European constitutional texts; without providing the model with the actual constitutional texts as context, we analyzed its capability to recognize and articulate on the main RD-relevant topics of the European constitutions. Second, more importantly, we defined a two-stage process of summarization, i.e., within- and across-country, for each of the RD-relevant topics, based on different choices of prompt engineering and incremental summary update. Overall, we assessed the model's ability to produce meaningful summaries across the constitutions of different countries with the ultimate objective of capturing a transnational context to lead towards a hypothetical unified constitution for all European countries according to their citizens' rights and duties.

[1] `https://platform.openai.com/docs/models/gpt-3-5`

## 2 Data

We retrieved the European constitutions from `www.constituteproject.org` [4], which provides free and public access to constitutions of world countries. Each of the constitutions was originally annotated according to a three-level hierarchy of topics. We narrowed down on the first-level topic "Rights and Duties" (RD), filtering in European countries only (43 in total), and extracted all the portions of English-written constitutions labeled with RD-relevant topics. These topics are 114 and organized in 9 second-level topics, or *macro-topics*. We will use symbols $t$ and $M$ to denote a topic and a macro-topic, respectively, and $T_M$ to denote the subset of topics from a macro-topic $M$. Table 1 provides the complete list of RD-relevant topics.

The data used in this study was previously described in [5] for purposes of explorative and similarity analysis of the European constitutions. Following their lead, we also carried out a text preprocessing step of *anonymization*, aimed to debias the analysis from country-specific terms, which were replaced by generic identifiers for different entities such as "geo-political European entity" in place of country names, "European people" for inhabitants, "person" for names of persons (e.g., royals, office secretaries), "organization", "location", etc.; all legal references were replaced with a special token.

## 3 Methodology

As previously mentioned, our model choice is `gpt-3.5-turbo` (for short, GPT-3.5T), which was accessed through OpenAI API based on authenticated requests.

Table 1: RD-relevant topics and their descriptions, grouped by the 9 macro-topics concerning Rights & Duties

| Topic id | Topic name |
|---|---|
| | (M1) **Physical Integrity Rights** |
| slave | Prohibition of slavery |
| cruelty | Prohibition of cruel treatment |
| cappun | Prohibition of capital punishment |
| life | Right to life |
| torture | Prohibition of torture |
| corppun | Prohibition of corporal punishment |
| | (M2) **Social Rights** |
| edcomp | Compulsory education |
| env | Protection of environment |
| finsup3 | State support for the disabled |
| safework | Right to safe work environment |
| provwork | Right to work |
| conright | Protection of consumers |
| childwrk | Limits on employment of children |
| standliv | Right to reasonable standard of living |
| edfree | Free education |
| finsup1 | State support for the elderly |
| finsup4 | State support for children |
| remuner | Right to just remuneration |
| shelter | Right to shelter |
| strike | Right to strike |
| health | Right to health care |
| achighed | Access to higher education |
| leisure | Right to rest and leisure |
| jointrde | Right to join trade unions |
| finsup2 | State support for the unemployed |
| water | Right to water |
| scifree | Right to enjoy the benefits of science |
| | (M3) **Economic Rights** |
| busines | Right to establish a business |
| intprop | Provisions for intellectual property |
| transfer | Right to transfer property |
| occupate | Right to choose occupation |
| proprght | Right to own property |
| exprop | Protection from expropriation |
| freecomp | Right to competitive marketplace |
| | (M4) **Citizen Duties** |
| milserv | Duty to serve in the military |
| taxes | Duty to pay taxes |
| work | Duty to work |
| | (M5) **General Duties** |
| abide | Duty to obey the constitution |
| binding | Binding effect of const rights |
| | (M6) **Civil and Political Rights** |
| assoc | Freedom of association |
| express | Freedom of expression |
| voteun | Claim of universal suffrage |
| assem | Freedom of assembly |
| fndfam | Right to found a family |
| freemove | Freedom of movement |
| libel | Right to protect one's reputation |
| nomil | Right to conscientious objection |
| civmar | Provision for civil marriage |
| freerel | Freedom of religion |
| arms | Right to bear arms |
| citren | Right to renounce citizenship |
| dignity | Human dignity |
| marriage | Regulation of marriage |
| childpro | Rights of children |
| acfree | Right to academic freedom |
| infoacc | Right to information |
| press | Freedom of press |
| debtors | Rights of debtors |
| devlpers | Right to development of personality |
| overthrw | Right to overthrow government |
| petition | Right of petition |
| privacy | Right to privacy |
| opinion | Freedom of opinion/thought/conscience |
| | (M7) **Legal Procedural Rights** |
| prerel | Right to pre-trial release |
| juvenile | Privileges for juveniles in criminal process |
| presinoc | Presumption of innocence in trials |
| wolaw | Principle of no punishment without law |
| miranda | Protection from self-incrimination |
| jury | Jury trials required |
| excrim | Extradition procedure |
| amparo | Right to amparo |
| fairtri | Right to fair trial |
| falseimp | Protection from false imprisonment |
| couns | Right to counsel |
| dueproc | Guarantee of due process |
| habcorp | Protection from unjustified restraint |
| appeal | Right to appeal judicial decisions |
| speedtri | Right to speedy trial |
| examwit | Right to examine evidence/witnesses |
| trilang | Trial in native language of accused |
| expost | Protection from ex post facto laws |
| vicright | Protection of victim's rights |
| doubjep | Prohibition of double jeopardy |
| pubtri | Right to public trial |
| evidence | Regulation of evidence collection |
| | (M8) **Enforcement** |
| hr | Human rights commission |
| amparo | Right to amparo |
| ombuds | Ombudsman |
| inalrght | Inalienable rights |
| | (M9) **Minority Rights** |
| equalgr2 | Equality regardless of nationality |
| equalgr13 | Equality regardless of financial status |
| equalgr5 | Equality regardless of language |
| indpolgr1 | Indigenous right to vote |
| indpolgr6 | Indigenous right to self governance |
| equalgr11 | Equality regardless of creed or belief |
| equalgr16 | Equality regardless of parentage |
| equalgr8 | Equality regardless of age |
| equalgr3 | Equality regardless of origin |
| equalgr10 | Equality regardless of skin color |
| rightres | Restrictions on rights of groups |
| equalgr1 | Equality regardless of gender |
| equalgr4 | Equality regardless of race |
| socclas | Mentions of social class |
| selfdet | Right to self determination |
| equal | General guarantee of equality |
| equalgr15 | Equality regardless of political party |
| equalgr12 | Equality regardless of social status |
| asylum | Protection of stateless persons |
| indpolgr2 | Indigenous right to representation |
| cultrght | Right to culture |
| matequal | Provision for matrimonial equality |
| equalgr6 | Equality regardless of religion |
| equalgr9 | Equality for persons with disabilities |
| indcit | Citizenship of indigenous groups |
| equalgr7 | Equality regardless of sexual orientation |

To maximally leverage the model's ability for the task of abstractive summarization, we resorted to its '16k' release, which allows us to accommodate the length of the constitution portions pertinent to an RD-relevant topic.

To address our first research question (**RQ1**), we delved into the generative model's awareness of the themes woven into the European constitutions. For this purpose, we asked GPT-3.5T to describe as many topics as possible about each particular macro-topic (selected from our dataset's topic-hierarchy), according to its knowledge of the European constitutions. In this stage, no portion of an European country's constitution was provided in prompts to GPT-3.5T. Specifically, we provided `gpt-3.5-turbo-4k` with the following prompt for each macro-topic $M$:

> Describe the topic <M> based on what is written in the constitutions of European countries. Provide a list of as many arguments as possible that are related to this topic, with the associated description. The list must be structured as a number, followed by the argument name, followed by the description.

Given a macro-topic $M$ to ask GPT3.5T about, each item in the topic list returned by the model was compared to the actual subset of topics for $M$ (i.e., $T_M$), in order to find the best matching in terms of semantic similarity. To this aim, we computed the cosine similarity between the sentence embedding of a model topic and the sentence embedding of an actual topic $t \in T_M$, which were produced by all-mpnet-base-v1,[2] a highly effective Sentence-Transformer [11].

Concerning **RQ2**, we devised a two-stage summarization task, which was applied to each macro-topic $M$. In Stage 1, we focused on each topic $t \in T_M$ at a time, then we selected all portions of a country's constitution that were originally labeled with $t$, with the purpose of generating a topic-specific summary for that country. To this aim, we asked GPT3.5T to first extract keywords from the text (i.e., set of constitution portions for that country) then to build a summary upon those keywords. Given a country $C$ and a topic $t$, for Stage 1 we provided the model with the following prompt:

> Given the text TEXT(C,t), detect keywords in the text and generate a compressed version of it based on the keywords. All keywords must appear in the compressed version of the text. Show in output the keywords and the compressed text in the following format:
> Keywords: <list-of-keywords>
> Compressed Text: <compressed-text>
> where <list-of-keywords> is the list of identified keywords and <compressed-text> is the generated compressed text.

In Stage 2, we requested the model to create a unified constitution summary for $t$ based on the country-level summaries acquired in Stage 1. Here we adopted an incremental approach, starting with the summary of largest constitution-portion and requesting one step at a time to incorporate the next country-level summary, in decreasing order of size of the original country portion-set. More specifically, we provided GPT3.5T with the current summary and the next country-level summary as a concatenated text and asked the model to produce a new summary by avoiding redundant information while retaining the keywords identified in Stage 1. Specifically, for Stage 2, we used the following prompt:

> Given the text:
> Current-TopicLevel-Summary
> Current-CountryLevel-Summary
> Detect redundant information into the text and reformulate it in order to have all information without redundancies. Note that you must keep the following keywords: list-of-keywords-current-topiclevel-summary, list-of-keywords-countrylevel-summary
> Return only the final complete text, in the following format:
> Final Text: <final-text>
> where <final-text> is the generated text.

Note that a "full-batch" approach could not be used without losing information as alternative to our defined incremental approach, due to imposed token-length limitations of GPT models. Since we strive to maintain each final topic-level summary as coherent as possible with the corresponding original texts in the constitutions, we set the GPT3.5T *temperature* to 0. Also, we assigned a *role* to the model, instructing it to act as *"summarization expert"*: by means of this directive, we actually enhanced the model's ability to effectively extract key aspects from the text and construct a summary that ensures the inclusion of essential information.

[2] https://www.sbert.net/docs/pretrained_models.html

**Evaluation criteria.** Our selected assessment criteria were mainly designed to compare a summary and its source. To this purpose, we used both *non-LLM-based* and *LLM-based* criteria.

The former group includes character-level **compression ratio** ($CR$); **novelty** ($N(I,S) = |S \setminus I|/|S|$), **Jaccard** ($J(I,S) = |S \cap I|/|S \cup I|$) and **Dice** ($D(I,S) = 2|S \cap I|/(|S| + |I|)$) coefficients over the sets of words (excluding stopwords) of input text $I$ and summary $S$; cosine-based *lexical similarity* between **TF.IDF** vectors of source/summary obtained after stopword removal and stemming; **Rouge** metrics [7] *R1*, *R2*, *RL*, and *RLSum*.

The LLM-based criteria include the cosine-based *semantic similarity* between **S-BERT** embeddings of a text/summary, based on 512-token-sized `all-mpnet-base-v1` LLM; **BLANC** [14] scores which involve BERT to estimate the quality of a summary without a reference summary, using *BLANC-help* (BH) and *BLANC-tune* (BT) (cf. Appendix A.2); a **qualitative evaluation** by asking GPT-3.5T to rate a summary on a scale 1(worst)-5(best) according to five criteria [1]: **Informative** - a summary is informative if it encapsulates the crucial details from the source, offering a precise and concise presentation; **Quality** - a summary has an high quality if it is understandable and comprehensible; **Coherence** - a summary is coherent if it demonstrates a sound structure and organization; **Attributable** - all the information in the summary are attributable to the source; **Overall Preference** - the summary should succinctly, logically, and coherently convey the primary ideas presented in the source.

It is worth noticing that we also carried out an evaluation stage based on human-generated summaries. To this purpose, since there are no publicly available reference summaries for (the RD-relevant portions of) the EU constitutions to date, we were assisted by legal professionals, with confidence on constitutional topics, who manually created summaries for each of the RD-relevant topics in M1. Note that, for the application of the non-LLM-based criteria, the input text $I$ refers to a reference summary.

## 4 Results

**RQ1 evaluation.** Table 2(left) reports on similarity results between the actual topics and the ones known to GPT3.5T. For each macro-topic $M$, the table shows the minimum, maximum, and mean similarity over all best matchings between a topic in $M$ and a topic generated by the model when asked about $M$. We observe relatively tight ranges of values over mid-high regimes of cosine similarity. This gives evidence of the model's knowledge about the existing RD topics, and its high confidence about the actual meanings within the constitutions. This is further confirmed by the descriptions of the generated topics and best-matching actual topics, reported in Appendix A.1.

Table 2: (left) Similarity between generated topics and best-matching actual topics, (right) Summarization results

| M | Most-sim/topic | Least-sim/topic | Mean |
|---|---|---|---|
| M1 | 0.778/slave | 0.667/corppun | 0.717 |
| M2 | 0.763/remuner | 0.554/childwrk | 0.637 |
| M3 | 0.708/intprop | 0.556/transfer | 0.636 |
| M4 | 0.875/taxes | 0.613/work | 0.730 |
| M5 | 0.766/abide | 0.533/abide | 0.646 |
| M6 | 0.867/privacy | 0.496/civmar | 0.669 |
| M7 | 0.855/speedtri | 0.449/excrim | 0.701 |
| M8 | 0.709/ombuds | 0.545/amparo | 0.569 |
| M9 | 0.716/indpolgr2 | 0.482/equalgr8 | 0.587 |

| M | #C | CR (%) | R1 | R2 | RL | RLSum | N (%) | J | D | S-BERT | TF.IDF | BH | BT |
|---|---|---|---|---|---|---|---|---|---|---|---|---|---|
| M1 | 28 | 45.856 | 0.571 | 0.460 | 0.474 | 0.526 | 6.48 | 0.696 | 0.816 | 0.877 | 0.823 | 0.390 | 0.263 |
| M2 | 24 | 57.250 | 0.638 | 0.506 | 0.559 | 0.592 | 7.53 | 0.640 | 0.772 | 0.909 | 0.904 | 0.290 | 0.296 |
| M3 | 30 | 48.776 | 0.599 | 0.486 | 0.543 | 0.560 | 4.72 | 0.639 | 0.771 | 0.864 | 0.902 | 0.241 | 0.289 |
| M4 | 17 | 65.227 | 0.728 | 0.571 | 0.641 | 0.669 | 7.85 | 0.718 | 0.835 | 0.922 | 0.908 | 0.338 | 0.377 |
| M5 | 16 | 70.981 | 0.779 | 0.612 | 0.723 | 0.723 | 4.57 | 0.738 | 0.848 | 0.938 | 0.941 | 0.316 | 0.361 |
| M6 | 27 | 49.611 | 0.586 | 0.468 | 0.538 | 0.554 | 8.99 | 0.587 | 0.726 | 0.880 | 0.901 | 0.245 | 0.297 |
| M7 | 22 | 56.701 | 0.670 | 0.570 | 0.616 | 0.634 | 5.22 | 0.705 | 0.819 | 0.888 | 0.918 | 0.282 | 0.317 |
| M8 | 15 | 65.124 | 0.717 | 0.634 | 0.677 | 0.688 | 3.89 | 0.733 | 0.829 | 0.796 | 0.884 | 0.331 | 0.374 |
| M9 | 18 | 46.224 | 0.573 | 0.476 | 0.528 | 0.541 | 5.41 | 0.612 | 0.746 | 0.872 | 0.858 | 0.225 | 0.247 |

Since the model was not given instructions on the number of topics to identify for each macro-topic, we also inspected if the model over-/underestimated the number of actual topics. By denoting with $(n, n')_M$ the actual ($n$) and generated ($n'$) topic numbers for a given macro-topic $M$, we found $(6, 15)_{M1}$, $(21, 14)_{M2}$, $(7, 15)_{M3}$, $(3, 15)_{M4}$, $(2, 15)_{M5}$, $(24, 20)_{M6}$, $(22, 18)_{M7}$, $(4, 15)_{M8}$, $(26, 15)_{M9}$: note that, although 4 out of 9 times the model underestimated the topic number, there was an overall marked opposite tendency, with 53 overestimated topics.

**RQ2 evaluation.** Table 2(right) shows the summarization performance scores averaged over all topics in each macro-topic. (In Appendix A.3, Table 4 contains the full topic-level results, whereas Figure 3 provides topic-level details for all criteria.) We observe that the per-macro-topic compression ratio varies from 46% to 71%, with a tendency to increase as the average number of involved countries (#C) decreases; note also that the latter, being lower than 43 for all macro-topics, generally indicates that only a subset of European constitutions contain portions that are RD-relevant.

Novelty of summaries is relatively low yet non-negligible (from 4% to 9%). Being within mid-high regimes, Jaccard and Dice similarity coefficients show a strong word-set sharing between the generated summaries and the original constitution-portions, which also couples with a very high lexical TF.IDF-vector-based similarity and semantic S-BERT-embedding-based similarity. In addition, ROUGE metrics are good indicators as well; in this respect, note that they tend to be higher compared to the domain-general abstractive/extractive summarization leaderboard scores.[3] Analogously, the observe BLANC scores (BT and BH) are generally higher than the typical threshold of 0.3 for indicating a good summary. This is likely due to the constrain we imposed to the model in generating summaries based on terms extracted from the original text, thus remaining as much faithful as possible to the constitution texts, which should be regarded as a desired requirement in any legal application of these summaries.

*Qualitative evaluation.* Figure 1 shows the distribution of rates provided by GPT-3.5T on the summaries according to the criteria of informativeness, quality, coherence, attributability, and overall preference. Note that this stage of evaluation was carried out independently from the summary generation process, i.e., GPT-3.5T was asked to carry out a qualitative evaluation on summaries it was unaware of. From the figure, we observe generally high values (above 4 out of 5) for all criteria, on each macro-topic, which means that summaries were evaluated as precise, concise, and attributable w.r.t. the source text, as well as coherent and understandable.

[3] `https://paperswithcode.com/task/abstractive-text-summarization`

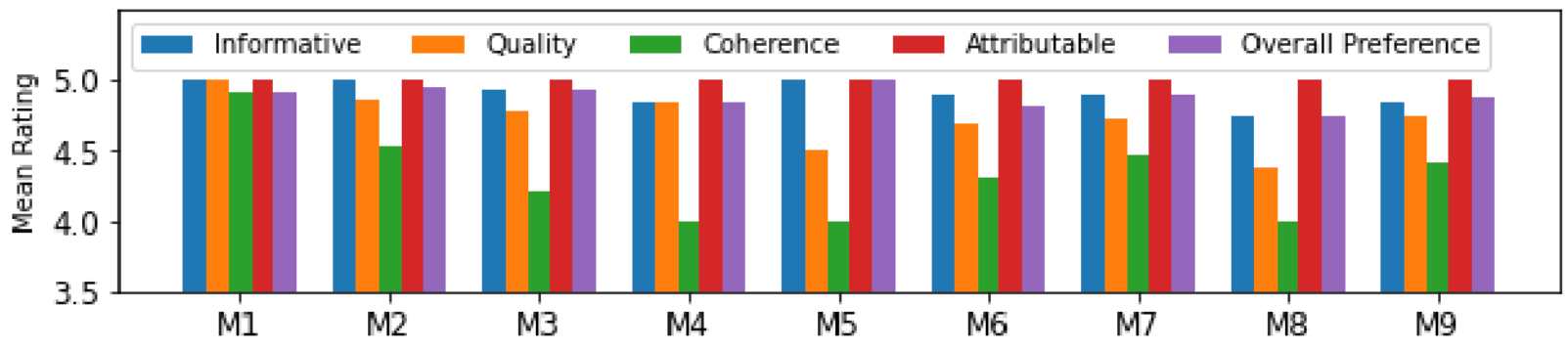

Fig. 1: Qualitative summary evaluation by GPT3.5T

Table 3: Evaluation on human-based reference summaries for macro-topic M1

| **Topic id** | **CR (%)** | **R1** | **R2** | **RL** | **RLsum** | **N (%)** | **J** | **D** | S-BERT | TF-IDF |
|---|---|---|---|---|---|---|---|---|---|---|
| slave | 105.81 | 0.693 | 0.378 | 0.246 | 0.423 | 31.64 | 0.531 | 0.693 | 0.916 | 0.781 |
| cruelty | 140.61 | 0.664 | 0.403 | 0.256 | 0.470 | 33.55 | 0.539 | 0.701 | 0.902 | 0.763 |
| cappun | 107.40 | 0.745 | 0.471 | 0.445 | 0.460 | 25.86 | 0.614 | 0.761 | 0.943 | 0.761 |
| life | 72.21 | 0.641 | 0.393 | 0.235 | 0.462 | 27.87 | 0.444 | 0.615 | 0.920 | 0.733 |
| torture | 138.81 | 0.673 | 0.404 | 0.267 | 0.496 | 40.46 | 0.506 | 0.672 | 0.939 | 0.655 |
| corppun | 117.01 | 0.643 | 0.360 | 0.296 | 0.383 | 40.35 | 0.459 | 0.630 | 0.936 | 0.446 |

*Human-based evaluation.* Table 3 shows results obtained by comparing the generated summaries and the ones manually crafted for the macro-topic M1. A few interesting remarks stand out, also comparatively to the first row of Table 2. First, the compression ratio values indicate that the generated and human summaries are mostly quite aligned in terms of length. They have very high semantic similarity (always above 0.9), whereas lexical similarity is still high but lower than the results in Table 2. This also couples with the relatively lower ROUGE scores except R1, which would suggest that the GPT3.5T summaries differ more in terms of syntactic structure and paraphrasing style, despite essential keywords (unigrams) might be more preserved, when compared to human-provided summaries than to the source texts. This is also confirmed by the much higher novelty, meaning that the generated summaries tend to include a significant amount of terms missing in the human summaries.

## 5 Discussion and Conclusions

We explored the use of OpenAI GPT-3.5 language model for understanding and summarizing European constitutional texts that are relevant to citizens' rights and duties. Results have shown an adequate level of knowledge by GPT3.5T about RD topics, and its ability of generating concise, informative, coherent and attributable summaries w.r.t. the EU countries' constitution sources.

Our research in this study represents a first step towards assessing the implications of utilizing generative LLMs for understanding and summarizing European Constitutions that address universally significant topics, such as those pertaining to the rights and duties of citizens. We believe this can pave the way for further developments that aims to support a variety of tasks oriented to a more transparent, integrated and cooperative transnational legal landscape. Nonetheless, our work has a number of limitations which are described next.

1. As noted in the Introduction, there exist various alternatives to our model choice not only looking at the OpenAI GPT family (e.g., GPT-4) or other commercially-licensed models but also to solutions based on Open LLMs (e.g., Llama, Mistral). Either types of LLMs should be taken into account to get a much clearer picture of the capabilities of such models in understanding and summarizing European constitutions, thus providing effective support for computational tasks on European law.

2. We believe that incorporating various European law resources (e.g., Official Journal of the European Union, European case-law, summaries of legislation, etc.) might strongly enhance the capability of the model of reasoning about European constitutions and related law data.

3. Exploiting reference summaries is essential for an in-depth assessment of summarization models, therefore we have carried out a preliminary evaluation based on summaries created by experts. However, given the lack of reference summarization for EU constitutions, our analysis was limited to one of the RD macro-topics only. Thus, a more comprehensive investigation is needed.

4. Country-focused evaluation is missing. It would be interesting to understand, especially in the stage of across-country summarization, whether and to what extent one or more countries could lead a "primary" role in building an overall encompassing summary for each (macro-)topic. Moreover, further investigation would be needed to delve into the generated summaries' quality aspects that closely deal with the legal sublanguage used in the European constitutions as well as with those (country-specific) terms that were anonymized.

5. Our study in this work is focused on European countries' constitutions and their relations with the rights and duties theme. Nonetheless, the proposed methodology can straightforwardly be extended to other countries/continents and/or thematic subjects in the constitutional context.

6. Potential risks: Although our model was prompted equally for all countries, aspects related to fairness, equity and legal coherence of summaries w.r.t. the involved countries have not been checked.

**Ethical remark:** The authors declare that any finding that might be drawn from the analysis provided in this work should be intended to support (legal) decision-making and not to replace the human specialists.

**Acknowledgements:** AT is partly funded by the PNRR Future AI Research (FAIR) project (H23C22000860006, spoke 9). This funder had no role in data collection and analysis, decision to publish, or preparation of the manuscript.

## A Appendix

### A.1 RQ1 best matchings

For each macro-topic, we report the best-matching pair of topic descriptions: on the bottom, the description generated by GPT3.5T, on the top, the one listed in ConstituteProject; the prefix in each description corresponds to the topic name.

*Macro-topic M1*:

Prohibition of slavery: Forbids the practice of owning or otherwise exploiting human beings. May still allow for compulsory labor as a punishment for a crime, and/or compulsory service in the military in time of war.
Prohibition of slavery and forced labor: Constitutions may include provisions that explicitly prohibit slavery, servitude, or forced labor. This protects individuals from being subjected to any form of forced labor or exploitation.

*Macro-topic M2*:

Right to just remuneration: Grants individuals the right to just remuneration or fair/equal payment for work. May also regulate wages including the establishment of a minimum or living wage.
Right to fair remuneration: This argument ensures that individuals have the right to fair remuneration for their work, including equal pay for equal work. It protects individuals from unfair wage practices and ensures that they receive a decent income to support themselves and their families.

*Macro-topic M3*:

Provisions for intellectual property: Gives individuals and businesses the exclusive right to protect and profit from their creative works.
Right to cultural and intellectual property: Some constitutions recognize the right to protect and benefit from one's cultural and intellectual creations, promoting innovation, creativity, and economic incentives for artists, inventors, and creators.

*Macro-topic M4*:

Duty to pay taxes: Requires all citizens to contribute to the expenses of the state through taxes. This is usually a general statement and may still allow for variation on the basis of income.
Duty to pay taxes: Citizens have a duty to pay taxes to contribute to the financing of public services and the functioning of the state. This duty is essential for the provision of public goods and services.

*Macro-topic M5*:

Duty to obey the constitution: Requires citizens or public officers to obey the constitution. This primarily applies to individuals in cases where the constitution includes a statement of the duties of citizens or holders of public office.
Maintenance of public order and security: Constitutions typically impose a duty on the state to maintain public order and security within its territory. This duty includes protecting citizens from crime, ensuring the functioning of the justice system, and maintaining law and order.

*Macro-topic M6*:

Right to privacy: Grants individuals protection against government intrusion into their property, papers, person, or communications without proper authorization. Also protects against such intrusions by private individuals.
Right to privacy: This right protects individuals' privacy, including their personal and family life, home, and correspondence, from arbitrary interference by the state or other individuals

*Macro-topic M7*:

Right to speedy trial: Grants individuals charged with an offense the right to be brought to trial without any undue delay.
Right to speedy trial: This right ensures that individuals are not subjected to undue delays in the legal process. It aims to prevent prolonged pre-trial detention and to ensure that justice is delivered in a timely manner.

*Macro-topic M8*:

Ombudsman: A public official tasked with investigating improper or illegal actions on the part of government agencies which may infringe on the rights of individuals. Enables individuals to bring complaints about the actions of the government.
Constitutional ombudsman: Some European countries have constitutional ombudsmen or similar institutions that are responsible for protecting citizens' rights and overseeing the enforcement of the constitution. These independent bodies investigate complaints, mediate disputes, and promote good governance.

*Macro-topic M9*:

*Indigenous right to representation*: Grants indigenous peoples the right to representation in the central government. This might, for example, involve a seat quota in the legislature.
*Representation and participation*: Minority rights include the right to representation and participation in decision-making processes at various levels of government. This can involve reserved seats or quotas to ensure minority groups have a voice in legislative bodies or other institutions.

### A.2 Implementation details

We used Python implementations available on NLTK for stemmming,[4] Scikit-learn for TF.IDF modeling,[5] HuggingFace for ROUGE,[6] and GitHub for BLANC.[7]

Concerning BLANC, this measures the efficacy of a summary in supporting a LLM in performing the Cloze task (masked token prediction) on a document. This measure is proposed in two versions [14]: **BLANC-Help**, which concatenates the summary to each sentence within the document, in order for the summary to be of assistance during inference; **BLANC-Tune**, which employs the summary for the fine-tuning phase, before analyzing the document. In both cases, the measure is calculated in terms of token unmasking accuracy, more specifically the difference in accuracy with and without the use of the summary. Note that BLANC has a limit on manageable tokens (512 tokens, as it employs BERT), which led to the truncation of summaries that exceeded this threshold; this occurred for 74 out of 114 topics in our case.

We employed the OpenAI APIs to prompt GPT-3.5T models, which implies the following costs: $0.0015/1K tokens for input and $0.002/1K tokens for output as regards `gpt-3.5-turbo-4k`, \$0.003/1K tokens for input and \$0.004/1K tokens for output as regards `gpt-3.5-turbo-16k`. The total cost for this research was about \$72, which includes various prompting trials.

[4] https://www.nltk.org/howto/stem.html
[5] https://scikit-learn.org/stable/modules/generated/sklearn.feature_extraction.text.TfidfVectorizer.html
[6] https://huggingface.co/spaces/evaluate-metric/rouge
[7] https://github.com/PrimerAI/blanc

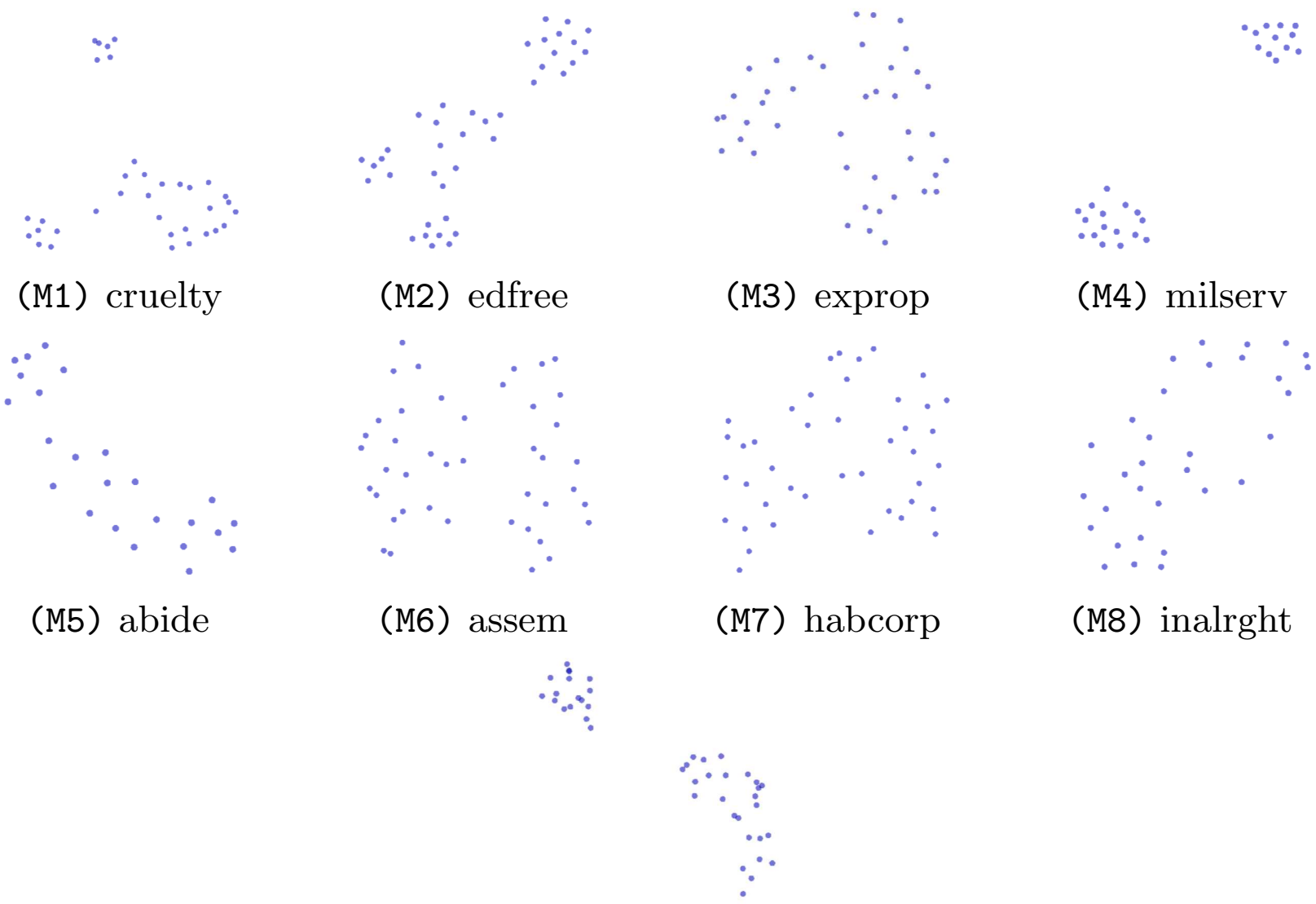

(M1) cruelty (M2) edfree (M3) exprop (M4) milserv

(M5) abide (M6) assem (M7) habcorp (M8) inalrght

(M9) equal

Fig. 2: For each macro-topic, the UMAP visualization of the country-level summary embeddings related to the topic with the highest number of participating countries.

### A.3 Additional Results

*Country-level summary representations.* We also investigated how the country-level summaries were consistent with each other in relation to the specific topic. In this respect, we leverage Uniform Manifold Approximation and Projection (UMAP) technique [10] to visualize the country-specific summary embeddings generated through S-BERT at Stage 1. Figure 2 shows for each macro-topic, the topic with the highest number of involved countries. We notice cohesive groups of country embeddings yet varying degree of separation among the countries across the topics.

*Compression ratio.* Table 4 reports the topic-level results for each metric. We notice that there are cases of excessively high or too mild compression ratio. A very high CR values, e.g., topic *standliv*, can be attributed to the model's failure to fully eliminate redundant concepts, causing the generated text to be unsuitable as a summary, or due to the imposition to preserve all the information (e.g. in the case of the topic *amparo* the CR value is 100.25%, this topic is covered

Table 4: Performance criteria for all topics

| Topic id | #C | CR (%) | R1 | R2 | RL | RLsum | N (%) | J | D | S-BERT | TF-IDF | BH | BT |
|---|---|---|---|---|---|---|---|---|---|---|---|---|---|
| slave | 29 | 39.723 | 0.537 | 0.436 | 0.418 | 0.504 | 5.650 | 0.628 | 0.771 | 0.800 | 0.812 | 0.266 | 0.239 |
| cruelty | 25 | 50.773 | 0.642 | 0.554 | 0.548 | 0.621 | 5.806 | 0.764 | 0.866 | 0.880 | 0.874 | 0.371 | 0.301 |
| cappun | 32 | 31.905 | 0.464 | 0.347 | 0.349 | 0.408 | 6.897 | 0.667 | 0.800 | 0.821 | 0.694 | 0.407 | 0.191 |
| life | 34 | 32.498 | 0.463 | 0.366 | 0.362 | 0.431 | 4.098 | 0.525 | 0.688 | 0.858 | 0.888 | 0.329 | 0.243 |
| torture | 34 | 43.875 | 0.568 | 0.462 | 0.512 | 0.529 | 7.634 | 0.807 | 0.893 | 0.933 | 0.823 | 0.456 | 0.285 |
| corppun | 5 | 76.365 | 0.752 | 0.597 | 0.655 | 0.662 | 8.772 | 0.788 | 0.881 | 0.971 | 0.846 | 0.511 | 0.319 |
| edcomp | 31 | 40.039 | 0.540 | 0.413 | 0.471 | 0.509 | 2.703 | 0.647 | 0.785 | 0.916 | 0.877 | 0.323 | 0.244 |
| env | 38 | 34.129 | 0.480 | 0.398 | 0.405 | 0.451 | 4.365 | 0.491 | 0.658 | 0.866 | 0.932 | 0.124 | 0.215 |
| finsup3 | 31 | 31.333 | 0.442 | 0.351 | 0.383 | 0.414 | 9.140 | 0.424 | 0.595 | 0.891 | 0.842 | 0.181 | 0.183 |
| safework | 17 | 53.275 | 0.625 | 0.498 | 0.581 | 0.581 | 6.504 | 0.685 | 0.813 | 0.915 | 0.884 | 0.323 | 0.277 |
| provwork | 32 | 64.184 | 0.712 | 0.540 | 0.572 | 0.652 | 8.000 | 0.687 | 0.814 | 0.904 | 0.897 | 0.255 | 0.321 |
| conright | 12 | 69.310 | 0.740 | 0.551 | 0.665 | 0.665 | 4.545 | 0.808 | 0.894 | 0.957 | 0.892 | 0.416 | 0.358 |
| childwrk | 17 | 48.122 | 0.605 | 0.485 | 0.512 | 0.562 | 7.843 | 0.588 | 0.740 | 0.956 | 0.876 | 0.309 | 0.259 |
| standliv | 20 | 117.418 | 0.780 | 0.585 | 0.623 | 0.731 | 4.831 | 0.801 | 0.889 | 0.938 | 0.909 | 0.255 | 0.367 |
| edfree | 39 | 48.110 | 0.602 | 0.484 | 0.530 | 0.572 | 12.438 | 0.570 | 0.726 | 0.943 | 0.940 | 0.245 | 0.245 |
| finsup1 | 26 | 68.413 | 0.768 | 0.637 | 0.719 | 0.723 | 6.084 | 0.742 | 0.852 | 0.949 | 0.963 | 0.242 | 0.367 |
| finsup4 | 24 | 53.355 | 0.652 | 0.518 | 0.590 | 0.605 | 6.529 | 0.630 | 0.773 | 0.804 | 0.913 | 0.222 | 0.313 |
| remuner | 22 | 69.758 | 0.750 | 0.600 | 0.673 | 0.693 | 8.333 | 0.691 | 0.817 | 0.951 | 0.951 | 0.256 | 0.311 |
| shelter | 16 | 58.365 | 0.678 | 0.518 | 0.523 | 0.633 | 7.729 | 0.710 | 0.830 | 0.839 | 0.884 | 0.276 | 0.323 |
| strike | 28 | 45.619 | 0.570 | 0.456 | 0.522 | 0.525 | 8.759 | 0.601 | 0.751 | 0.961 | 0.935 | 0.329 | 0.332 |
| health | 33 | 60.292 | 0.736 | 0.655 | 0.685 | 0.716 | 3.361 | 0.691 | 0.817 | 0.896 | 0.975 | 0.278 | 0.338 |
| achighed | 20 | 68.056 | 0.763 | 0.628 | 0.705 | 0.724 | 3.529 | 0.820 | 0.901 | 0.972 | 0.955 | 0.317 | 0.293 |
| leisure | 22 | 39.768 | 0.545 | 0.435 | 0.478 | 0.513 | 7.071 | 0.529 | 0.692 | 0.928 | 0.879 | 0.349 | 0.286 |
| jointrde | 38 | 19.786 | 0.311 | 0.254 | 0.234 | 0.305 | 4.380 | 0.354 | 0.523 | 0.775 | 0.900 | 0.284 | 0.164 |
| finsup2 | 27 | 57.268 | 0.686 | 0.544 | 0.611 | 0.628 | 4.603 | 0.673 | 0.804 | 0.874 | 0.925 | 0.256 | 0.310 |
| water | 3 | 91.043 | 0.833 | 0.720 | 0.778 | 0.778 | 10.000 | 0.831 | 0.908 | 0.949 | 0.873 | 0.522 | 0.507 |
| scifree | 5 | 64.609 | 0.579 | 0.356 | 0.489 | 0.462 | 27.273 | 0.464 | 0.634 | 0.909 | 0.791 | 0.321 | 0.214 |
| busines | 25 | 60.023 | 0.709 | 0.581 | 0.653 | 0.663 | 6.276 | 0.718 | 0.836 | 0.873 | 0.923 | 0.333 | 0.383 |
| intprop | 23 | 57.029 | 0.682 | 0.600 | 0.646 | 0.668 | 2.280 | 0.781 | 0.877 | 0.813 | 0.878 | 0.115 | 0.254 |
| transfer | 23 | 57.566 | 0.675 | 0.532 | 0.611 | 0.638 | 5.051 | 0.746 | 0.855 | 0.959 | 0.900 | 0.316 | 0.312 |
| occupate | 33 | 38.111 | 0.501 | 0.383 | 0.407 | 0.431 | 6.207 | 0.560 | 0.718 | 0.958 | 0.857 | 0.363 | 0.269 |
| proprght | 40 | 28.843 | 0.422 | 0.331 | 0.379 | 0.388 | 5.263 | 0.422 | 0.594 | 0.687 | 0.924 | 0.128 | 0.183 |
| exprop | 42 | 33.330 | 0.475 | 0.403 | 0.445 | 0.452 | 1.984 | 0.492 | 0.660 | 0.806 | 0.945 | 0.158 | 0.204 |
| freecomp | 21 | 66.533 | 0.727 | 0.571 | 0.659 | 0.679 | 5.991 | 0.753 | 0.859 | 0.952 | 0.889 | 0.272 | 0.416 |
| milserv | 28 | 62.349 | 0.717 | 0.572 | 0.625 | 0.666 | 10.549 | 0.673 | 0.805 | 0.950 | 0.955 | 0.254 | 0.380 |
| taxes | 18 | 65.903 | 0.763 | 0.628 | 0.701 | 0.738 | 2.703 | 0.788 | 0.882 | 0.967 | 0.928 | 0.405 | 0.419 |
| work | 5 | 67.430 | 0.704 | 0.514 | 0.597 | 0.602 | 10.294 | 0.693 | 0.819 | 0.850 | 0.840 | 0.355 | 0.331 |
| abide | 24 | 62.595 | 0.737 | 0.605 | 0.683 | 0.681 | 5.039 | 0.698 | 0.822 | 0.912 | 0.941 | 0.210 | 0.334 |
| binding | 8 | 79.368 | 0.821 | 0.619 | 0.763 | 0.766 | 4.110 | 0.778 | 0.875 | 0.964 | 0.941 | 0.422 | 0.389 |
| assoc | 41 | 39.191 | 0.532 | 0.434 | 0.473 | 0.509 | 7.469 | 0.497 | 0.664 | 0.896 | 0.956 | 0.215 | 0.255 |
| express | 42 | 40.061 | 0.544 | 0.455 | 0.497 | 0.511 | 5.502 | 0.604 | 0.753 | 0.848 | 0.945 | 0.105 | 0.211 |
| voteun | 34 | 64.709 | 0.762 | 0.691 | 0.734 | 0.742 | 1.460 | 0.813 | 0.897 | 0.870 | 0.959 | 0.140 | 0.376 |
| assem | 43 | 37.243 | 0.509 | 0.429 | 0.468 | 0.483 | 8.854 | 0.526 | 0.689 | 0.872 | 0.924 | 0.230 | 0.258 |
| fndfam | 25 | 64.381 | 0.752 | 0.660 | 0.720 | 0.730 | 6.329 | 0.779 | 0.876 | 0.961 | 0.933 | 0.364 | 0.404 |
| freemove | 35 | 26.691 | 0.389 | 0.317 | 0.346 | 0.372 | 6.283 | 0.476 | 0.645 | 0.856 | 0.882 | 0.261 | 0.254 |
| libel | 26 | 29.107 | 0.428 | 0.344 | 0.382 | 0.395 | 2.128 | 0.354 | 0.523 | 0.739 | 0.832 | 0.279 | 0.212 |
| nomil | 23 | 54.219 | 0.666 | 0.548 | 0.595 | 0.634 | 6.154 | 0.667 | 0.800 | 0.933 | 0.951 | 0.315 | 0.311 |
| civmar | 5 | 87.967 | 0.698 | 0.513 | 0.593 | 0.646 | 18.750 | 0.605 | 0.754 | 0.950 | 0.791 | 0.310 | 0.379 |
| freerel | 43 | 11.130 | 0.195 | 0.155 | 0.164 | 0.185 | 4.167 | 0.247 | 0.397 | 0.864 | 0.822 | 0.176 | 0.107 |
| arms | 1 | 45.528 | 0.444 | 0.000 | 0.444 | 0.444 | 50.000 | 0.250 | 0.400 | 0.729 | 0.777 | 0.083 | 0.417 |
| citren | 12 | 41.749 | 0.508 | 0.418 | 0.439 | 0.471 | 9.524 | 0.623 | 0.768 | 0.957 | 0.871 | 0.500 | 0.444 |
| dignity | 35 | 62.550 | 0.736 | 0.614 | 0.685 | 0.685 | 4.696 | 0.697 | 0.821 | 0.923 | 0.975 | 0.173 | 0.312 |
| marriage | 24 | 56.742 | 0.711 | 0.620 | 0.673 | 0.695 | 3.333 | 0.751 | 0.858 | 0.903 | 0.962 | 0.345 | 0.330 |
| childpro | 32 | 50.087 | 0.623 | 0.485 | 0.542 | 0.578 | 5.785 | 0.663 | 0.797 | 0.753 | 0.944 | 0.190 | 0.247 |
| acfree | 28 | 72.705 | 0.815 | 0.720 | 0.791 | 0.791 | 4.188 | 0.792 | 0.884 | 0.931 | 0.970 | 0.285 | 0.371 |
| infoacc | 28 | 46.450 | 0.593 | 0.480 | 0.535 | 0.560 | 8.127 | 0.558 | 0.716 | 0.921 | 0.939 | 0.144 | 0.213 |
| press | 38 | 30.197 | 0.457 | 0.409 | 0.428 | 0.443 | 2.941 | 0.431 | 0.602 | 0.998 | 0.909 | 0.148 | 0.206 |
| debtors | 3 | 52.941 | 0.479 | 0.252 | 0.364 | 0.364 | 29.412 | 0.500 | 0.667 | 0.909 | 0.629 | 0.162 | 0.135 |
| devlpers | 13 | 61.121 | 0.717 | 0.575 | 0.682 | 0.669 | 3.185 | 0.764 | 0.866 | 0.949 | 0.923 | 0.333 | 0.375 |
| overthrw | 6 | 89.522 | 0.858 | 0.755 | 0.843 | 0.848 | 10.000 | 0.797 | 0.887 | 0.984 | 0.948 | 0.562 | 0.508 |
| petition | 33 | 61.888 | 0.718 | 0.583 | 0.660 | 0.662 | 7.203 | 0.668 | 0.801 | 0.914 | 0.952 | 0.186 | 0.346 |
| privacy | 41 | 31.410 | 0.455 | 0.375 | 0.406 | 0.434 | 4.516 | 0.530 | 0.693 | 0.617 | 0.909 | 0.210 | 0.263 |
| opinion | 38 | 33.069 | 0.474 | 0.406 | 0.439 | 0.455 | 5.652 | 0.501 | 0.668 | 0.841 | 0.929 | 0.167 | 0.193 |
| prerel | 16 | 69.006 | 0.770 | 0.629 | 0.690 | 0.708 | 7.738 | 0.738 | 0.849 | 0.928 | 0.939 | 0.275 | 0.319 |
| juvenile | 9 | 51.623 | 0.650 | 0.540 | 0.528 | 0.611 | 2.128 | 0.724 | 0.840 | 0.949 | 0.873 | 0.451 | 0.409 |
| presinoc | 32 | 51.769 | 0.634 | 0.509 | 0.556 | 0.571 | 4.800 | 0.708 | 0.829 | 0.926 | 0.923 | 0.389 | 0.312 |
| wolaw | 35 | 65.580 | 0.783 | 0.695 | 0.740 | 0.738 | 2.326 | 0.840 | 0.913 | 0.911 | 0.983 | 0.196 | 0.344 |
| miranda | 19 | 50.586 | 0.611 | 0.480 | 0.543 | 0.589 | 7.222 | 0.668 | 0.801 | 0.727 | 0.773 | 0.209 | 0.231 |
| jury | 19 | 61.550 | 0.744 | 0.680 | 0.729 | 0.723 | 1.176 | 0.737 | 0.848 | 0.840 | 0.941 | 0.350 | 0.382 |
| excrim | 26 | 55.054 | 0.677 | 0.580 | 0.636 | 0.636 | 6.757 | 0.706 | 0.828 | 0.953 | 0.977 | 0.321 | 0.353 |
| amparo | 1 | 100.250 | 1.000 | 1.000 | 1.000 | 1.000 | 0.000 | 1.000 | 1.000 | 1.000 | 1.000 | 0.646 | 0.646 |
| fairtri | 27 | 48.677 | 0.641 | 0.567 | 0.605 | 0.618 | 3.984 | 0.722 | 0.838 | 0.887 | 0.942 | 0.239 | 0.348 |
| falseimp | 24 | 50.873 | 0.628 | 0.501 | 0.564 | 0.587 | 5.759 | 0.655 | 0.791 | 0.784 | 0.862 | 0.247 | 0.249 |
| couns | 29 | 55.554 | 0.679 | 0.581 | 0.618 | 0.644 | 1.794 | 0.718 | 0.836 | 0.911 | 0.958 | 0.235 | 0.340 |
| dueproc | 6 | 94.937 | 0.970 | 0.955 | 0.970 | 0.952 | 1.266 | 0.951 | 0.975 | 0.997 | 0.991 | 0.425 | 0.517 |
| habcorp | 42 | 15.275 | 0.263 | 0.220 | 0.218 | 0.249 | 6.383 | 0.290 | 0.450 | 0.761 | 0.892 | 0.120 | 0.160 |
| appeal | 29 | 55.118 | 0.679 | 0.568 | 0.622 | 0.632 | 4.235 | 0.724 | 0.840 | 0.861 | 0.943 | 0.148 | 0.268 |
| speedtri | 27 | 42.769 | 0.562 | 0.446 | 0.482 | 0.517 | 9.548 | 0.577 | 0.732 | 0.871 | 0.892 | 0.227 | 0.273 |
| examwit | 9 | 67.422 | 0.724 | 0.567 | 0.676 | 0.688 | 9.278 | 0.721 | 0.838 | 0.946 | 0.895 | 0.318 | 0.216 |
| trilang | 18 | 55.578 | 0.639 | 0.498 | 0.570 | 0.584 | 8.939 | 0.688 | 0.815 | 0.943 | 0.927 | 0.244 | 0.307 |
| expost | 32 | 51.515 | 0.657 | 0.562 | 0.578 | 0.628 | 6.832 | 0.698 | 0.822 | 0.907 | 0.974 | 0.222 | 0.283 |
| vicright | 5 | 80.641 | 0.798 | 0.605 | 0.719 | 0.735 | 9.639 | 0.806 | 0.893 | 0.961 | 0.919 | 0.396 | 0.325 |
| doubjep | 17 | 41.464 | 0.539 | 0.433 | 0.494 | 0.501 | 5.882 | 0.584 | 0.737 | 0.807 | 0.775 | 0.158 | 0.163 |
| pubtri | 36 | 35.398 | 0.506 | 0.436 | 0.465 | 0.470 | 4.706 | 0.572 | 0.728 | 0.908 | 0.927 | 0.275 | 0.291 |
| evidence | 32 | 46.777 | 0.597 | 0.495 | 0.545 | 0.561 | 4.337 | 0.680 | 0.809 | 0.758 | 0.896 | 0.121 | 0.233 |
| hr | 2 | 81.758 | 0.829 | 0.651 | 0.749 | 0.749 | 6.383 | 0.830 | 0.907 | 0.949 | 0.883 | 0.412 | 0.375 |
| ombuds | 26 | 23.880 | 0.357 | 0.286 | 0.315 | 0.339 | 4.926 | 0.448 | 0.619 | 0.661 | 0.812 | 0.075 | 0.136 |
| inalrght | 32 | 54.610 | 0.684 | 0.599 | 0.644 | 0.666 | 4.255 | 0.655 | 0.792 | 0.575 | 0.842 | 0.193 | 0.310 |
| equalgr2 | 13 | 73.825 | 0.803 | 0.684 | 0.760 | 0.763 | 7.843 | 0.750 | 0.857 | 0.937 | 0.948 | 0.244 | 0.359 |
| equalgr13 | 20 | 50.437 | 0.669 | 0.576 | 0.592 | 0.637 | 0.826 | 0.759 | 0.863 | 0.916 | 0.943 | 0.242 | 0.284 |
| equalgr5 | 24 | 56.295 | 0.714 | 0.627 | 0.674 | 0.693 | 3.247 | 0.764 | 0.866 | 0.954 | 0.959 | 0.279 | 0.260 |
| indpolgr1 | 1 | 20.798 | 0.272 | 0.142 | 0.254 | 0.233 | 5.195 | 0.332 | 0.498 | 0.809 | 0.598 | 0.117 | 0.084 |
| indpolgr6 | 4 | 18.614 | 0.281 | 0.193 | 0.243 | 0.246 | 6.849 | 0.320 | 0.485 | 0.808 | 0.767 | 0.164 | 0.129 |
| equalgr11 | 21 | 51.770 | 0.626 | 0.539 | 0.599 | 0.596 | 5.957 | 0.602 | 0.752 | 0.824 | 0.875 | 0.170 | 0.228 |
| equalgr16 | 14 | 44.510 | 0.633 | 0.535 | 0.592 | 0.596 | 2.174 | 0.687 | 0.814 | 0.959 | 0.914 | 0.344 | 0.237 |
| equalgr8 | 6 | 58.895 | 0.617 | 0.427 | 0.588 | 0.577 | 7.921 | 0.669 | 0.802 | 0.925 | 0.804 | 0.220 | 0.229 |
| equalgr3 | 22 | 52.926 | 0.628 | 0.539 | 0.600 | 0.605 | 8.163 | 0.634 | 0.776 | 0.838 | 0.880 | 0.160 | 0.230 |
| equalgr10 | 13 | 41.040 | 0.531 | 0.448 | 0.500 | 0.511 | 9.036 | 0.559 | 0.717 | 0.843 | 0.825 | 0.192 | 0.215 |
| rightres | 6 | 70.774 | 0.733 | 0.584 | 0.661 | 0.673 | 8.029 | 0.700 | 0.824 | 0.888 | 0.859 | 0.426 | 0.338 |
| equalgr1 | 38 | 46.728 | 0.603 | 0.521 | 0.566 | 0.575 | 4.096 | 0.585 | 0.739 | 0.823 | 0.936 | 0.092 | 0.244 |
| equalgr4 | 30 | 45.361 | 0.592 | 0.509 | 0.557 | 0.563 | 6.564 | 0.610 | 0.757 | 0.772 | 0.924 | 0.152 | 0.261 |
| socclas | 12 | 87.199 | 0.909 | 0.846 | 0.892 | 0.894 | 2.622 | 0.858 | 0.924 | 0.863 | 0.966 | 0.210 | 0.496 |
| selfdet | 14 | 48.189 | 0.625 | 0.506 | 0.576 | 0.577 | 2.961 | 0.637 | 0.778 | 0.890 | 0.902 | 0.242 | 0.285 |
| equal | 43 | 49.019 | 0.644 | 0.576 | 0.613 | 0.620 | 2.730 | 0.679 | 0.809 | 0.847 | 0.970 | 0.159 | 0.299 |
| equalgr15 | 22 | 47.409 | 0.590 | 0.492 | 0.536 | 0.554 | 7.391 | 0.617 | 0.763 | 0.863 | 0.893 | 0.155 | 0.230 |
| equalgr12 | 22 | 51.341 | 0.677 | 0.582 | 0.613 | 0.637 | 1.587 | 0.761 | 0.864 | 0.967 | 0.936 | 0.298 | 0.301 |
| asylum | 28 | 47.113 | 0.617 | 0.525 | 0.556 | 0.588 | 3.043 | 0.670 | 0.802 | 0.912 | 0.923 | 0.273 | 0.304 |
| indpolgr2 | 2 | 31.547 | 0.469 | 0.442 | 0.463 | 0.467 | 3.030 | 0.790 | 0.883 | 0.915 | 0.797 | 0.334 | 0.288 |
| cultrght | 35 | 26.830 | 0.404 | 0.350 | 0.375 | 0.387 | 2.121 | 0.394 | 0.566 | 0.792 | 0.869 | 0.176 | 0.180 |
| matequal | 20 | 55.574 | 0.624 | 0.501 | 0.543 | 0.568 | 5.634 | 0.698 | 0.822 | 0.944 | 0.891 | 0.402 | 0.357 |
| equalgr6 | 34 | 25.179 | 0.399 | 0.349 | 0.330 | 0.388 | 1.550 | 0.410 | 0.581 | 0.881 | 0.917 | 0.242 | 0.184 |
| equalgr9 | 10 | 56.885 | 0.666 | 0.492 | 0.533 | 0.590 | 9.000 | 0.728 | 0.843 | 0.959 | 0.883 | 0.309 | 0.213 |
| indcit | 2 | 12.567 | 0.179 | 0.114 | 0.162 | 0.166 | 8.000 | 0.223 | 0.365 | 0.700 | 0.482 | 0.079 | 0.071 |
| equalgr7 | 5 | 31.009 | 0.383 | 0.272 | 0.346 | 0.363 | 15.172 | 0.484 | 0.653 | 0.842 | 0.648 | 0.176 | 0.123 |

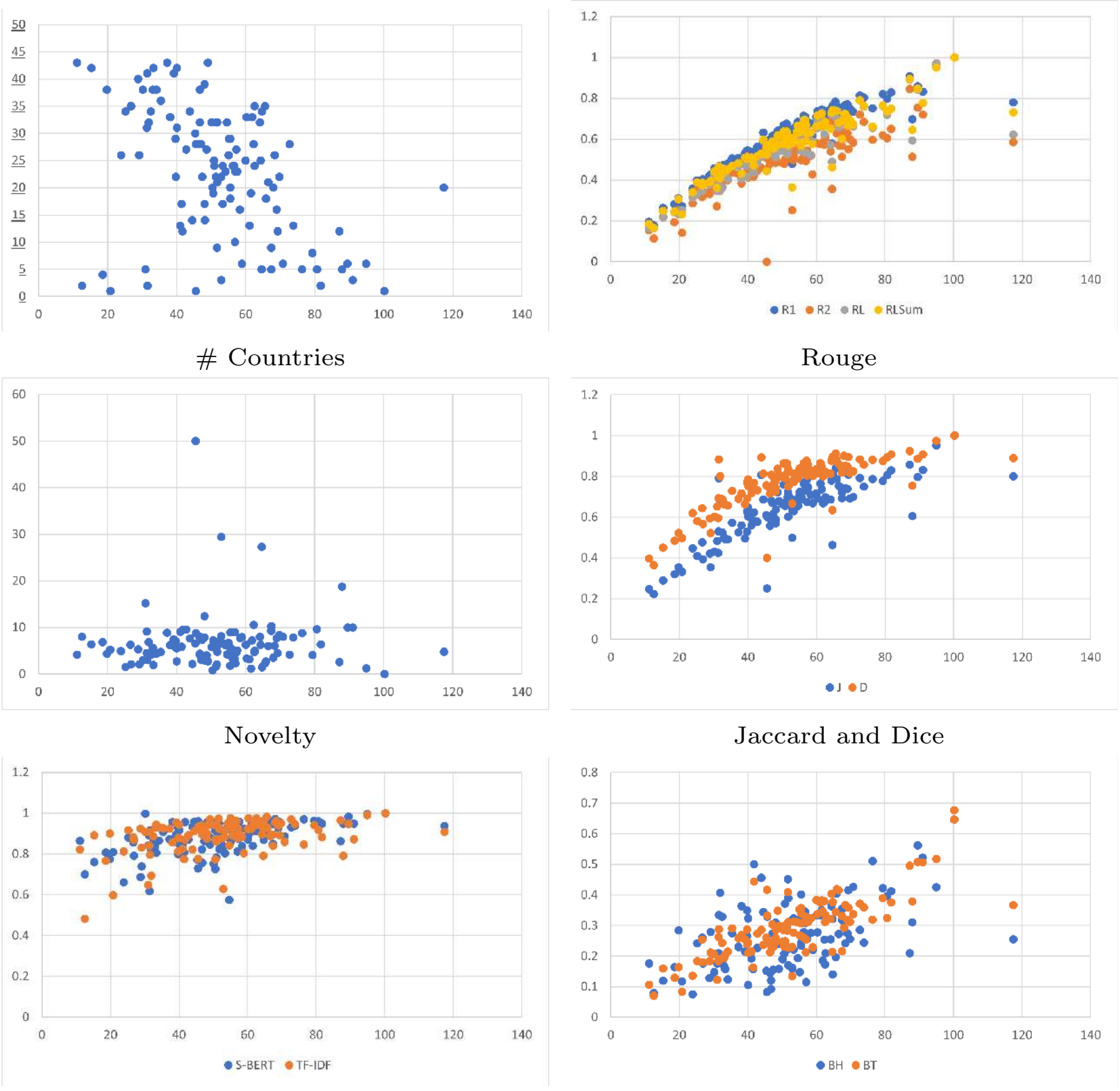

\# Countries | Rouge | Novelty | Jaccard and Dice | Text Similarity | Blanc

Fig. 3: Compression ratio (x-axis) versus each of the assessment criteria, for all RD-relevant topics

by only one country, so all the included information is preserved). On the other hand, too low CR values (e.g., in *freerel* and *indcit*) can be attributed to excessive redundancy of concepts among the involved countries, which should be absent in the summary, or to a significant loss of informative content. Fortunately, these extreme cases are few: out of 114 topics, only 9 topics have CR exceeding 80% and 12 topics have CR below 30%.

Figure 3 shows scatter plots to compare the compression ratio to each of the assessment criteria for all RD-relevant topics. It can be noted that the ROUGE, Jaccard, Dice and BLANC values increase as the compression ratio increases, which is not surprising. By contrast, there is an inverse correlation between compression ratio and number of countries, since the more countries are involved, the larger and more redundant the overall text of the topic of interest is likely to be. Novelty and textual similarity instead appear to be independent of the compression ratio.